%% file: main.tex
\documentclass[10pt,letterpaper]{article}
\ifdefined\pdfminorversion\pdfminorversion=7\fi
\usepackage[T1]{fontenc}
\usepackage[utf8]{inputenc}
\usepackage[letterpaper,top=0.83in,bottom=0.86in,left=0.96in,right=0.96in]{geometry}
\usepackage{amsmath,amssymb,newtxtext,newtxmath}
\usepackage{microtype,graphicx,booktabs,tabularx,array,enumitem,caption,placeins,multicol,float}
\usepackage[numbers,sort&compress]{natbib}
\usepackage{xcolor}
\definecolor{linktone}{HTML}{245465}
\usepackage[colorlinks=true,allcolors=linktone,pdfencoding=auto,psdextra,pdfusetitle]{hyperref}
\hypersetup{pdfauthor={Shenyan Guo, Yuming Huang, Yu Zhang}}
\usepackage{titlesec}
\titleformat{\section}{\large\bfseries}{\thesection}{0.65em}{}
\titleformat{\subsection}{\normalsize\bfseries}{\thesubsection}{0.65em}{}
\titlespacing*{\section}{0pt}{2ex plus 0.5ex minus 0.3ex}{0.9ex}
\titlespacing*{\subsection}{0pt}{1.65ex plus 0.4ex minus 0.2ex}{0.6ex}
\setlist{leftmargin=*,nosep,topsep=3pt}
\renewcommand{\arraystretch}{1.12}
\newcolumntype{Y}{>{\raggedright\arraybackslash}X}
\newcolumntype{P}[1]{>{\raggedright\arraybackslash}p{#1}}
\newcommand{\argmax}{\operatorname*{arg\,max}}
\newcommand{\stopact}{\textup{STOP}}
\newcommand{\runin}[1]{\par\noindent\textbf{#1.}\enspace}
\title{BeliefNav: Deferred-Commitment Beliefs with a Structured Decision Model for Recoverable Vision-and-Language Navigation}
\author{Shenyan Guo\textsuperscript{*\textdagger} \and Yuming Huang\textsuperscript{*} \and Yu Zhang}
\date{}
\makeatletter
\renewcommand{\@maketitle}{
  \begin{center}
  {\LARGE\bfseries \@title\par}
  \vspace{0.75em}
  {\large\renewcommand{\and}{\qquad}\@author\par}
  \vspace{0.35em}
  {\small Department of Robotics, University of Macau, Macau\par}
  {\footnotesize \textsuperscript{*}These authors contributed equally.\qquad
  \textsuperscript{\textdagger}Corresponding author.\par}
  \end{center}
}
\makeatother
\begin{document}
\maketitle
\input{sections/introduction}
\input{sections/related_work}
\input{sections/method}
\begingroup
\setlength{\intextsep}{8pt plus 2pt minus 2pt}
\input{sections/experiments}
\endgroup
\input{sections/discussion}
\FloatBarrier
\appendix
\setcounter{table}{0}
\renewcommand{\thetable}{A\arabic{table}}
\renewcommand{\theHtable}{appendix.\arabic{table}}
\numberwithin{equation}{section}
\input{appendix/settings}
\FloatBarrier
\input{appendix/baseline}
\FloatBarrier
\input{appendix/recovery}
\FloatBarrier
\input{appendix/audit}
\FloatBarrier
\begin{multicols}{2}[\section*{References}]
\begingroup
\renewcommand{\bibsection}{}
\fontsize{8}{9}\selectfont

\input{references.tex}
\endgroup
\end{multicols}
\end{document}

%% file: sections/introduction.tex
\begin{abstract}
An agent can see an instructed doorway before it reaches the correct turn. To keep that observation from prematurely settling instruction progress, BeliefNav maintains a joint belief over the current stage and the viewpoints at which the agent may have left the route, using a frozen navigation-trained backbone. At each viewpoint, a Gaussian mixture preserves disagreement between two assessments of the available waypoints, while a ray-guided comparison can revise their ranking or leave it unresolved. A neutral reference corrects for mass introduced by branching, and an adaptive gate selects viewpoint weights without a tuned cutoff. On the 100-episode R2R-CE subset, BeliefNav reaches 50.0\% success and 29.8\% SPL, an 18-percentage-point gain over visual reassessment with historical propagation. State-based control raises success by 4 percentage points with VLM logits and 12 with Jev. Within five actions of a deviation, BeliefNav returns to the reference route in 13 of 49 episodes, compared with 4 of 53 for history-based backtracking.
The code is available at \url{https://github.com/yuminhhuang/BeliefNav}.
\end{abstract}

\section{Introduction}
\label{sec:intro}
\begin{figure}[htbp]
\centering
\includegraphics[width=.68\linewidth]{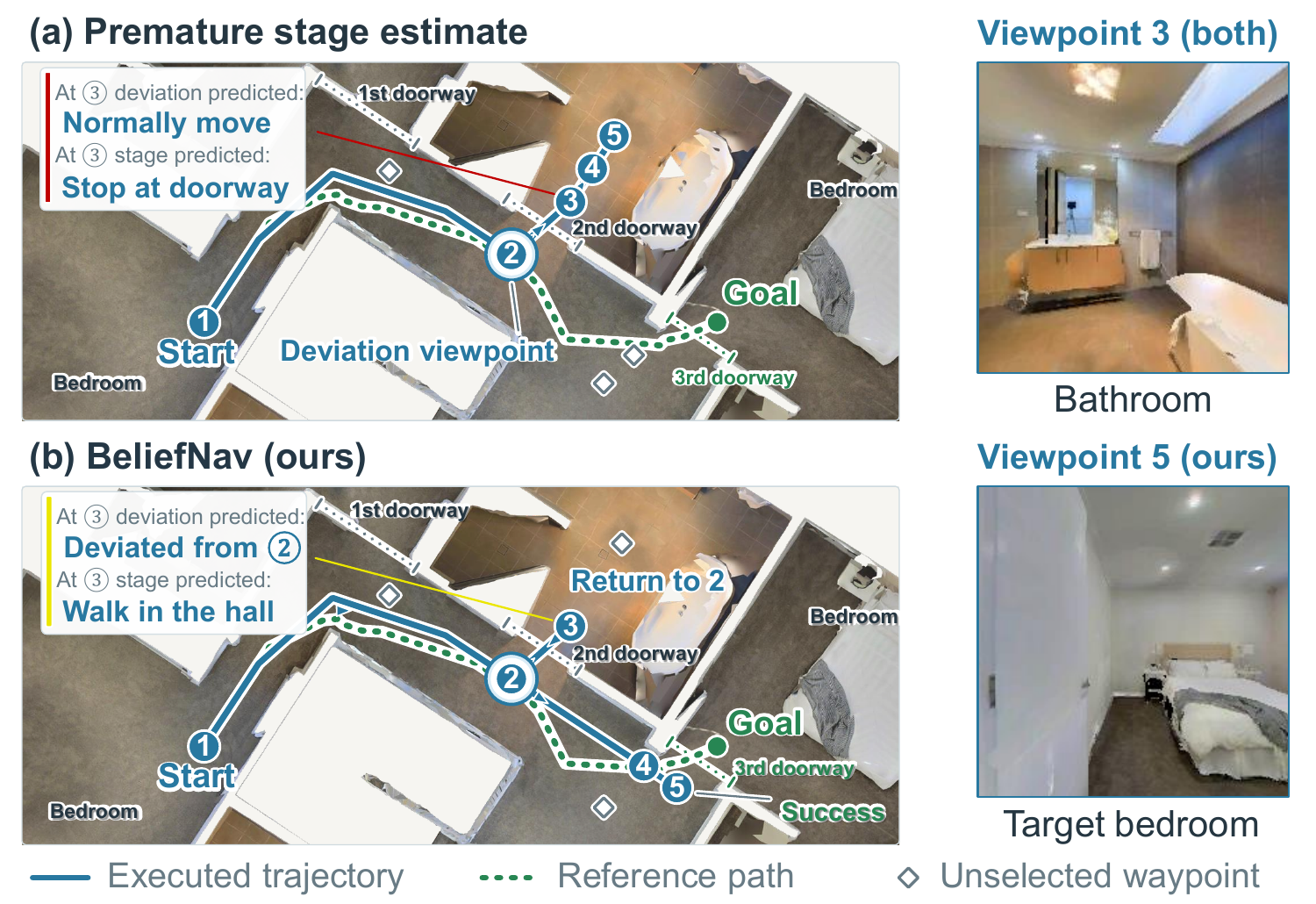}
\caption{\textbf{A visible doorway can precede the instructed turn.} The instruction reads \emph{“Leave the bedroom and take a right in the hall, and walk straight. Take a left into the third entry way on your left and enter the bedroom. Stop in the doorway to the bedroom.”} (a) The early stage estimate leads the agent farther from the route. (b) BeliefNav returns to viewpoint~2 and reaches the target bedroom.}
\label{fig:motivation}
\end{figure}

An instruction asks an agent to walk down a hall, take the third doorway on the left, and stop at the bedroom doorway. At viewpoint~3 in Fig.~\ref{fig:motivation}, a doorway is already visible, although the agent is still partway down the hall. Its stage estimate has advanced to the final clause. Under that interpretation the wrong doorway looks like a plausible continuation, giving the agent little reason to reconsider the turn.

Zero-shot navigation systems use history, topological maps, and sub-instruction constraints to track progress~\cite{zhou2024navgpt,chen2024mapgpt,chen2025canav}. Because the same doorway can match several stages, the motion between observations helps determine when that match should guide the next move.

BeliefNav keeps both the current stage and possible departure viewpoints uncertain. Two assessments of a viewpoint can disagree even after a comparison of their preferred moves. A neutral filter discounts support introduced by the opportunity to branch, so that evidence from an earlier viewpoint can guide the next decision.

We vary the semantic evaluator and controller independently, then compare waypoint choices under fixed observations and measure returns to the reference route in closed loop. This lets us distinguish a revised local ranking from a recovery completed by the moving agent.

%% file: sections/related_work.tex
\section{Related Work}
\label{sec:related}
\runin{Typed semantic decisions}
Jev returns a distribution over an application-defined, bounded answer space~\cite{typesafe2026api}. Visual Jev uses a shared visual context to answer separate questions about the same images~\cite{yu2026visual}. Such interfaces have been applied to service orchestration, memory, and model escalation~\cite{li2026edge,jiang2026memory,li2026judge}. In BeliefNav a typed decision is an observation tied to a viewpoint and a set of candidates. A pairwise question admits three answers, including an unresolved ordering. The controller can consume either a structured response or a distribution obtained from VLM logits.

\runin{Progress, memory, and planning in zero-shot VLN}
R2R and VLN-CE established instruction following in discrete and continuous environments~\cite{anderson2018vln,krantz2020vlnce}. NavGPT, DiscussNav, MapGPT, InstructNav, Open-Nav, and CA-Nav use navigation history, planning, or sub-instruction constraints to guide movement~\cite{zhou2024navgpt,long2024discuss,chen2024mapgpt,long2024instructnav,qiao2025opennav,chen2025canav}. We couple progress to a possible departure point so that an earlier viewpoint can regain influence over a later decision.

\runin{Progress estimation and recovery}
Self-Monitoring learns a progress signal alongside visual--textual grounding~\cite{ma2019selfmonitoring}. The Regretful Agent uses progress to choose between forward motion and backtracking~\cite{ma2019regretful}. HAMT encodes panoramic and action history~\cite{chen2021hamt}, and NavMorph models navigation dynamics with contextual memory~\cite{yao2025navmorph}. Here each recovery hypothesis retains the instruction stage at which the agent may have left the route.

\runin{Topology and waypoint selection}
Structured Scene Memory and DUET use maps for global planning~\cite{wang2021ssm,chen2022duet}. Waypoint prediction extends hierarchical navigation to continuous spaces~\cite{krantz2021waypoint}. ETPNav and BEVBert combine online topology with waypoint or local metric features~\cite{an2025etpnav,an2023bevbert}. HNR predicts future representations, DGNav adjusts graph density and connectivity, and ETP-R1 trains a graph policy with reinforcement fine-tuning~\cite{wang2024hnr,peng2026dgnav,ye2025etpr1}. We reuse the map, candidate proposals, and motion interface of ETP-R1. BeliefNav changes how observed viewpoints contribute to candidate selection.

\runin{Probabilistic state estimation}
Bayesian filtering and Markov localization retain alternative explanations as a robot moves~\cite{fox1999markov,thrun2005probabilistic}. We apply discrete forward filtering to instruction stage and departure viewpoint. The neutral filter supplies a reference for the probability mass induced by the transitions themselves.

%% file: sections/method.tex
\section{Method}
\label{sec:method}
\subsection{Problem Formulation and Decision Interface}
\label{sec:interface}
At step $t$ the agent has an instruction $I$, observations from visited viewpoints, and a history of executed motions. A frozen backbone maintains the topological map and proposes waypoints~\cite{ye2025etpr1}. Let $\mathcal H_t$ contain the visited viewpoints with eligible candidates and retained assessments. For $h\in\mathcal H_t$, $\mathcal G_h$ contains the currently selectable waypoints associated with its observed views. A waypoint can occur in several such sets and keeps its identity in their union $\mathcal G_t$. A viewpoint index identifies an observation step, so revisiting a physical position can yield a new index.

\begin{figure}[htbp]
\centering
\includegraphics[width=\linewidth]{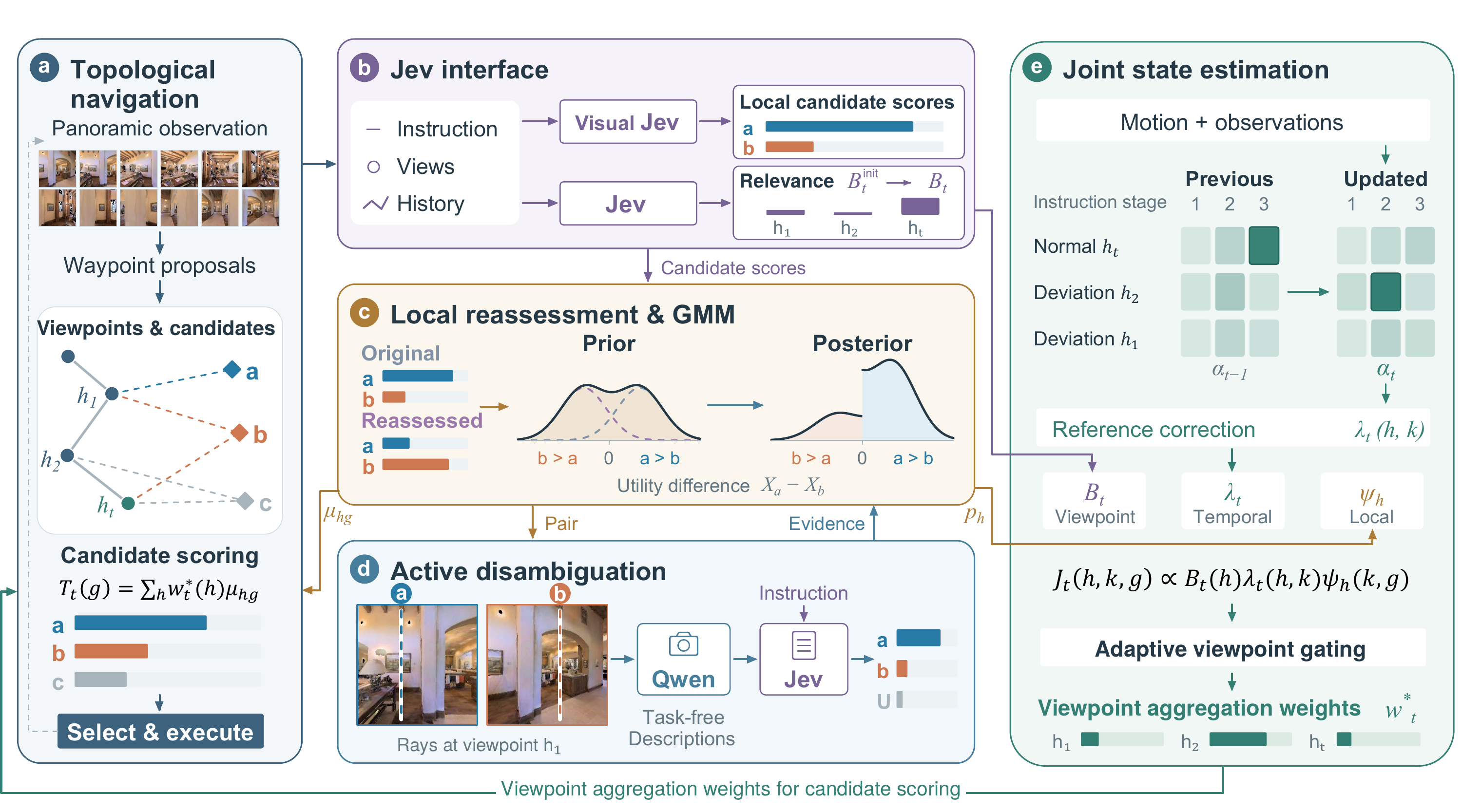}
\caption{\textbf{From observed viewpoints to a waypoint choice.} (a) The backbone supplies the map and proposals. (b) Visual Jev scores views and estimates initial relevance $B_t^{\rm init}$. Text-based Jev reassesses the available continuations to obtain $B_t$. (c) A mixture retains both local assessments. (d) A comparison of two bearings can favor either direction or remain unresolved. (e) The gate combines the progress--deviation belief with local utility distributions to choose the weights in Eq.~\eqref{eq:score}. Images and ray coordinates are recorded. Other values are schematic, and the heatmaps share a scale.}
\label{fig:framework}
\end{figure}

Each viewpoint supplies a mean utility $\mu_{hg}$ for moving toward candidate $g$. On a movement step we select
\begin{equation}
 T_t(g;w)=\sum_{h\in\mathcal H_t:\,g\in\mathcal G_h}w(h)\mu_{hg},
 \qquad g_t=\argmax_{g\in\mathcal G_t}T_t(g;w).
 \label{eq:score}
\end{equation}
The weights sum to one over viewpoints, and only viewpoints supporting $g$ contribute to its score. Figure~\ref{fig:framework} shows how we revise local utilities and select viewpoint weights. The backbone handles native \stopact{} before each movement decision.

\subsection{Gaussian Mixture over Local Waypoint Utilities}
\label{sec:mixture}
At viewpoint $h$, the semantic evaluator assigns each recorded camera view $v\in\mathcal V_h$ a score $p_h^{\rm cam}(v)$ using the full instruction and a fallible stage cue from the executed history. Projecting these scores onto candidate bearing $\theta_{hg}$ gives
\begin{equation}
 s_{hg}=\sum_{v\in\mathcal V_h}p_h^{\rm cam}(v)[\cos(\theta_{hg}-\theta_v)]_+,
 \qquad [x]_+=\max(x,0),
 \label{eq:projection}
\end{equation}
where $\theta_v$ is the camera heading. This projection has no normalization across candidates. Its value for an individual waypoint therefore stays fixed when another candidate is added to the same view.

A viewpoint weight scales all local utilities equally, preserving their ordering even when the stage cue favors the wrong passage. The initial relevance distribution $B_t^{\rm init}$ ranks viewpoints. We reassess up to two using the full instruction with the stage cue removed. Projecting the initial and reassessed camera scores through Eq.~\eqref{eq:projection} gives $s_h=(s_{hg})_{g\in\mathcal G_h}$ and $r_h=(r_{hg})_{g\in\mathcal G_h}$, respectively. All eligible viewpoints remain available for action selection.

Let $X_h$ be the vector of local utilities. When a compatible reassessment covers the current candidate set, the original and reassessed vectors define a Gaussian mixture~\cite{bishop2006prml},
\begin{equation}
 p_0(X_h)=\sum_{m\in\{s,r\}}\alpha_{hm}\mathcal N(X_h;\mu_{hm},\sigma^2\mathbf I),
 \qquad \mu_{hs}=s_h,\quad\mu_{hr}=r_h,\quad
 \alpha_{hs}=\alpha_{hr}=\tfrac12 .
 \label{eq:mixture}
\end{equation}
With a single assessment we use its Gaussian alone. We fix $\sigma^2=1/12$, the variance of a uniform score on $[0,1]$. Each draw shares one mixture component across all candidates, preserving disagreement over their ordering. The retained local posterior $p_h$ is $p_0$ or its update in Sec.~\ref{sec:pair}. It gives the mean $\mu_{hg}=\mathbb E_{p_h}[X_{hg}]$ for Eq.~\eqref{eq:score} and the local winning probabilities.

\subsection{Active Disambiguation}
\label{sec:pair}
\runin{Candidate-pair selection}
Two passages can fall within one camera view and receive similar projected scores. We use a ray-guided pairwise comparison~\cite{bajcsy1988active} when the current top candidate's estimated win probability is below $0.9$ (Eq.~\eqref{eq:outcome} with the current weights and utilities), or when the top candidate differs from the historical-propagation winner (Appendix~\ref{app:baseline}). The second global candidate is the historical winner when they disagree and the runner-up otherwise. Each candidate is assigned to the viewpoint making the largest contribution to its score. If the viewpoints differ, the controller uses their reassessed relevance $B_t$ (Sec.~\ref{sec:gate}). If they coincide at $h$, we choose a local incumbent $a$ and challenger $b$,
\begin{equation}
 a=\argmax_{g\in\mathcal G_h}\mu_{hg},\qquad
 b=\argmax_{g\in\mathcal G_h\setminus\{a\}}
 \sum_m\alpha_{hm}[\mu_{hmg}-\mu_{hma}]_+ .
 \label{eq:challenger}
\end{equation}
Here $\alpha_{hm}$ is the retained weight of component $m$, and $\mu_{hmg}$ is coordinate $g$ of its current mean at viewpoint $h$. We request at most one local comparison per decision and only if the challenger's positive advantage exceeds $10^{-12}$.

\runin{Ray-guided visual observation}
For each candidate, the representative associated camera view has the optical axis nearest its bearing, with ties broken by camera identity. For a $90^\circ$ horizontal field of view, the normalized horizontal image coordinate is
\begin{equation}
 x=\tfrac12-\tfrac12\tan(\theta_{hg}-\theta_v),\qquad
 |\theta_{hg}-\theta_v|<45^\circ .
 \label{eq:ray}
\end{equation}
Bearings increase counterclockwise from the viewpoint's forward axis. Each valid association receives a labeled vertical guide on its own observed image. Identical images with identical guides do not trigger a repeat comparison. A task-free Qwen observer~\cite{bai2025qwen25vl} describes the guide and its surroundings. Jev~\cite{typesafe2026api} compares the descriptions against the full instruction and history. It returns $(q_A,q_B,q_U)$ for preference for $a$, preference for $b$, or unresolved ordering. The three probabilities sum to one.

\runin{Posterior utility update}
The pairwise ranking likelihood is $\ell_+=q_A+q_U$ on $X_{ha}>X_{hb}$ and $\ell_-=q_B+q_U$ otherwise. Writing this piecewise factor as $\ell_q(X_h)$ gives $p_h(X_h)\propto p_0(X_h)\ell_q(X_h)$. Let $e_a,e_b\in\{0,1\}^{|\mathcal G_h|}$ be the one-hot vectors for candidates $a$ and $b$. Omitting the viewpoint subscript $h$ from the component means, define
\begin{equation}
 d_m=\frac{\mu_{ma}-\mu_{mb}}{\sqrt{2\sigma^2}},\qquad
 v=\sqrt{\sigma^2/2}\,(e_a-e_b).
 \label{eq:scale}
\end{equation}
With $\Phi$ and $\phi$ the standard normal CDF and density, the unnormalized component mass and first moment are
\begin{equation}
 Z_m=\ell_+\Phi(d_m)+\ell_-\Phi(-d_m),\qquad
 M_m=Z_m\mu_m+(\ell_+-\ell_-)\phi(d_m)v .
 \label{eq:moment}
\end{equation}
The posterior component weight is $\alpha_{hm}Z_m/\sum_j\alpha_{hj}Z_j$, and its conditional mean is $M_m/Z_m$. To estimate $L_h$ and $A_w$ in Secs.~\ref{sec:state}--\ref{sec:gate}, we draw each vector with one mixture component and one ranking halfspace shared across its coordinates. Equal preferences leave the distribution unchanged, including the fully unresolved response $q_U=1$. A new comparison replaces the preceding factor on the same assessment prior. This prevents repeated readings of one image from accumulating as independent evidence.

\subsection{Joint Estimation of Progress and Deviation}
\label{sec:state}
\runin{Joint state space and forward filtering}
A state $(a,k)$ describes either normal continuation $(\mathsf N,k)$ at instruction clause $k$ or departure from viewpoint $h$ at stage $k$, written $(h,k)$. Both the observation filter $\alpha$ and its neutral reference $\alpha^0$ start with mass $1/K$ at each of the $K$ normal stages and zero mass at departure anchors. The motion-dependent transition kernel $\mathcal T_t$ predicts the next state, which the observation factor $\ell_t$ updates as
\begin{equation}
 \alpha_t^-(a,k)=\sum_{a',j}\mathcal T_t(a,k\mid a',j)\alpha_{t-1}(a',j),
 \qquad \alpha_t(a,k)=\frac{\ell_t(a,k)\alpha_t^-(a,k)}
 {\sum_{a',j}\ell_t(a',j)\alpha_t^-(a',j)} .
 \label{eq:filter}
\end{equation}
A movement can open a departure branch when the agent turns or changes height, or when its starting viewpoint offers several directions or shows a doorway or passageway. Normal mass then splits equally between a new anchor and continuation. The continuing mass splits equally between staying and advancing if a fresh caption matches a feature of the next clause absent from the current one. Each departure anchor retains its stage. On returning to an anchor and leaving in a different direction, half its mass returns to normal. Anchor matching uses a numerical tolerance of $10^{-3}$\,m on the stored coordinates of discrete graph nodes.

\runin{Observation likelihood}
Deterministic clause segmentation gives features $\mathcal F_k$ from tokens, a fixed synonym map, and adjacent content-word pairs. Generic background nouns do not contribute alone. For a caption feature set $\mathcal O$, define
\begin{equation}
 \omega_f=1+\log\frac{K+1}{1+\#\{j:f\in\mathcal F_j\}},\qquad
 S(k,\mathcal O)=1+
 \frac{\sum_{f\in\mathcal F_k\cap\mathcal O}\omega_f}
      {\sum_{f\in\mathcal F_k}\omega_f}.
 \label{eq:features}
\end{equation}
An empty $\mathcal F_k$ gives $S=1$. Let $\mathcal D_t$ contain eligible fresh caption sets. The best matching fresh description supplies the normal-mode observation factor for each stage,
\begin{equation}
 R_t(k)=\max\big(\{1\}\cup\{S(k,\mathcal O):\mathcal O\in\mathcal D_t\}\big),\qquad
 f_t(k)=\frac{R_t(k)}{K^{-1}\sum_j R_t(j)},\qquad
 \ell_t(\mathsf N,k)=f_t(k),\quad \ell_t(h,k)=1 .
 \label{eq:emission}
\end{equation}
A repeated image is counted at its earliest observation. A panorama summary requires all its images to be fresh. Individual fresh views still contribute under partial overlap. Missing or stage-uniform support leaves the update neutral.

\runin{Neutral-reference bias correction}
Even with uninformative observations, repeated opportunities to initialize or recover a branch redistribute probability among viewpoints. The reference $\alpha^0$ follows the same transition gates under unit observation factors. Associate the current viewpoint with $a_h=\mathsf N$ and each earlier viewpoint with $a_h=h$. Assign zero reference mass to unrepresented modes and define
\begin{equation}
 r_t^0(a)=\sum_j\alpha_t^0(a,j),\qquad
 \lambda_t(h,k)=
 \begin{cases}
  \alpha_t(a_h,k)/r_t^0(a_h),&r_t^0(a_h)>0,\\
  1/K,&r_t^0(a_h)=0 .
 \end{cases}
 \label{eq:neutral}
\end{equation}
Every represented mode has positive reference mass. A navigable viewpoint whose departure branch was never initialized receives uniform stage support. An inactive recovery anchor remains in both temporal filters even when it has no navigable candidate. We retain the scale of $\lambda_t$ across stages so its sum can reflect evidence for that viewpoint.

\runin{Compatibility of stage, viewpoint, and candidate}
Write $R_h(k)$ for $R_t(k)$ from Eq.~\eqref{eq:emission} at the observation step indexed by $h$, $\mathcal V_{hg}$ for the distinct associated camera views of $g$, and $n_h=|\mathcal G_h|$. Candidate-specific caption support and local win probability give
\begin{align}
 L_h(g)&=\Pr_{X_h\sim p_h}\!\left[g=\argmax_{j\in\mathcal G_h}X_{hj}\right],&
 R_{hg}(k)&=\max\big(\{R_h(k)\}\cup\{S(k,\mathcal O_v):v\in\mathcal V_{hg}\}\big),
 \label{eq:local}\\
 C_{hkg}&=R_{hg}(k)/R_h(k),&
 \psi_h(k,g)&=
 \begin{cases}
 C_{hkg}\,n_hL_h(g),&g\in\mathcal G_h,\\
 0,&g\notin\mathcal G_h .
 \end{cases}
 \label{eq:psi}
\end{align}
The ratio $C_{hkg}$ replaces the viewpoint-level caption support where a candidate view provides a stronger match. With no additional support it equals one. The factor $n_hL_h(g)$ measures winning probability relative to uniform choice among the local candidates. We keep this scale when constructing the joint interpretation,
\begin{equation}
 W_t(h,k,g)=B_t(h)\lambda_t(h,k)\psi_h(k,g),\qquad
 J_t(h,k,g)=\frac{W_t(h,k,g)}
 {\sum_{r\in\mathcal H_t}\sum_{j=1}^{K}\sum_{u\in\mathcal G_r}W_t(r,j,u)} .
 \label{eq:joint}
\end{equation}
Equation~\eqref{eq:joint} combines reassessed viewpoint relevance $B_t$, corrected temporal support $\lambda_t$, and local compatibility $\psi_h$ into a product-of-experts distribution over supported candidates. Inactive anchors remain in the temporal filters.

\subsection{Final Action Selection by Adaptive Viewpoint Gating}
\label{sec:gate}
The belief-informed score $v^D$ pools stage support at each viewpoint and takes the strongest support for $g$. The prior-only score $v^R$ uses local win probabilities and viewpoint relevance, omitting the temporal factor $\lambda_t$ and caption factor $C_{hkg}$. With the same visibility constraint for both scores,
\begin{equation}
 v^D(g)=\max_{h\in\mathcal H_t:\,g\in\mathcal G_h}\sum_k J_t(h,k,g),\qquad
 v^R(g)=\max_{h\in\mathcal H_t:\,g\in\mathcal G_h} B_t(h)n_hL_h(g).
 \label{eq:targets}
\end{equation}
The target candidate distributions $q^D$ and $q^R$ place uniform mass on the exact maximizers of $v^D$ and $v^R$, respectively, and zero elsewhere.

The admissible weights are $\mathcal W=\{B_t,\delta_1,\ldots,\delta_H\}$, where $H=|\mathcal H_t|$ and $\delta_i\in\{0,1\}^{H}$ assigns all weight to viewpoint $i$. For each $w$, the local posteriors induce a distribution over winning actions,
\begin{equation}
 A_w(g)=\Pr\!\left[g=\argmax_{j\in\mathcal G_t}
       \sum_{h\in\mathcal H_t:\,j\in\mathcal G_h}w(h)X_{hj}\right].
 \label{eq:outcome}
\end{equation}
We estimate $L_h$ from 200,000 joint local draws per viewpoint and $A_w$ from 32,768 common draws per assignment. Draws are independent across viewpoints and share one mixture component within each viewpoint. Local numerical ties use the fixed candidate order.

\runin{Adaptive zero-hyperparameter minimax gate}
For either target distribution $q$, $A_w^\top q$ measures agreement with the modeled action. The gain over $B_t$ and the selected weights are
\begin{equation}
 \Delta(w;q)=(A_w-A_{B_t})^\top q,\qquad
 w_t^\star=\argmax_{w\in\mathcal W}\min\{\Delta(w;q^D),\Delta(w;q^R)\}.
 \label{eq:gate}
\end{equation}
$B_t$ has zero gain under both target distributions. A single viewpoint receives unit weight only for a strictly positive worst-case gain. If the best gain is zero, including a tie, the gate retains $B_t$. This rule sets the threshold from the baseline's own action distribution. Every full-system experiment uses Eq.~\eqref{eq:gate} followed by the posterior-mean action in Eq.~\eqref{eq:score}. A zero weight affects the current action only. Its viewpoint's belief remains available at later steps.

\runin{One movement decision}
\begin{enumerate}
 \item Rank viewpoints by $B_t^{\rm init}$ and reassess up to two using the full instruction.
 \item Construct the local mixtures and reassess viewpoint relevance. Given response probabilities $\rho_h$ for each eligible viewpoint and $\rho_U$ for an unresolved choice, set $B_t(h)=\rho_h+\rho_U B_t^{\rm init}(h)$.
 \item If eligible, compare competing bearings and condition their joint local utilities.
 \item Update the progress--deviation and neutral filters, then form $J_t$ from the retained posteriors, captions, motion, and $B_t$.
 \item Select $w_t^\star$ by Eq.~\eqref{eq:gate} and execute the highest-scoring waypoint in Eq.~\eqref{eq:score}.
\end{enumerate}

%% file: sections/experiments.tex
\section{Experiments}
\label{sec:experiments}
\subsection{Experimental Setup and Implementation Details}
\label{sec:setup}
\runin{Episodes and metrics}
We use Open-Nav's 100 R2R-CE validation-unseen episodes across ten scenes~\cite{qiao2025opennav,krantz2020vlnce}, with the released starts, goals, instructions, and reference routes. Following ETP-R1's standard panoramic discretization, initial headings align with the backbone's discrete viewing directions and are shared across internal comparisons. Each closed-loop configuration completes all 100 episodes using the same topology, a 15-step horizon including native \stopact{}, and a 3\,m final-position success radius. We report success (SR), oracle success (OSR), success weighted by path length (SPL), normalized dynamic time warping (nDTW)~\cite{ilharco2019ndtw}, and final navigation error (NE).

\runin{Frozen navigation and model roles}
ETP-R1 supplies the topological map, candidate proposals, low-level motion, collision handling, and native \stopact{}~\cite{ye2025etpr1}. Equation~\eqref{eq:score} selects movement independently of ETP-R1's learned movement scores. Visual Jev uses Qwen3-VL-4B-Instruct with its decision adapter for camera scores and initial viewpoint relevance~\cite{yu2026visual,bai2025qwen3vl}. Qwen2.5-VL-3B produces instruction-free descriptions of observed RGB images~\cite{bai2025qwen25vl}. Text-based Jev evaluates stages, pairwise preferences, and available route continuations~\cite{typesafe2026api}. We keep all models frozen and add no navigation training. The VLM evaluator disables the decision adapter and normalizes the base model's option-token logits at temperature one. Both evaluators share the task-free observer and receive identical visual and textual inputs when scoring the same decision.

\begin{figure}[htbp]
\centering
\includegraphics[width=\linewidth]{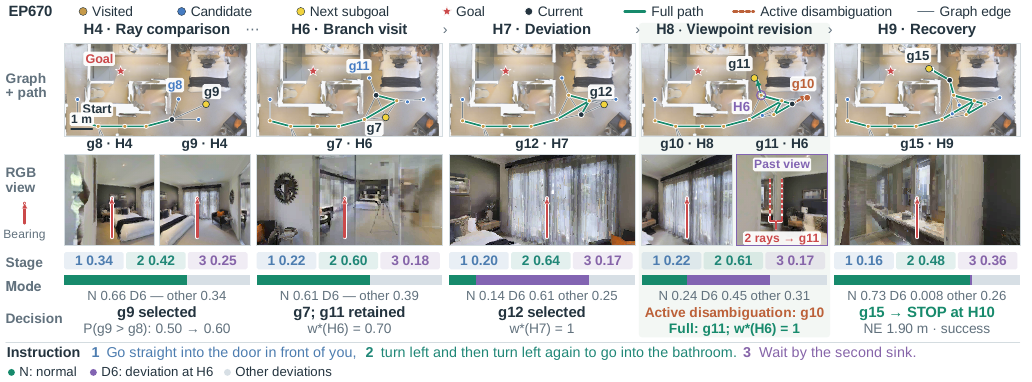}
\caption{\textbf{State revision in episode 670.} Columns align candidate views and bearings with the map, stage and mode probabilities, and chosen moves. High-level decisions are numbered from H0. At H8 two bearings support g11 through H6. Full BeliefNav and local active disambiguation finish 1.90 and 4.59\,m from the goal, respectively.}
\label{fig:timeline}
\end{figure}

\runin{Qualitative Case Study (Episode 670)}
At H4 in Fig.~\ref{fig:timeline}, the ray-guided comparison raises the preference for g9 over g8 from 0.50 to 0.60. Full BeliefNav and the local active-disambiguation variant take the same first eight actions (H0--H7). At H8 they have the same images and local utility moments but different viewpoint-relevance judgments. The full system's $B_t$ favors g10. Equation~\eqref{eq:gate} selects $w^\star(\mathrm{H7})=1$ at H7 and $w^\star(\mathrm{H6})=1$ at H8. The latter choice makes g11 the winning waypoint and takes the agent back through H6.

\subsection{Ablation and Cumulative Component Analysis}
\label{sec:cumulative}
The historical baseline already uses visual relevance to select viewpoints for reassessment (Appendix~\ref{app:baseline}). Substituting this relevance for propagation in the movement score raises SR from 32.0\% to 35.0\% (Table~\ref{tab:cumulative}). The stage cue gives 38.0\%. The local mixture, active comparison, and reassessed viewpoint relevance give 47.0\%. Joint estimation and gating raise SR to 50.0\%, while SPL falls from 31.1\% to 29.8\% and nDTW from 44.8\% to 41.9\%.

\begin{table}[htbp]
\centering\small
\caption{\textbf{Success and path quality as components are added.} Each row adds to the preceding row on the same 100 episodes. SR, OSR, SPL, and nDTW are percentages, NE is in meters, and Steps is the mean number of high-level decisions.}
\label{tab:cumulative}
\input{tables/cumulative}
\end{table}

\subsection{Backend-Agnostic Analysis of Scoring and State-Based Control}
\label{sec:crossed}
\runin{Closed-loop trajectories}
We cross two semantic evaluators with two controllers (Table~\ref{tab:crossed}). The score-based controller uses contextual stage cues and viewpoint-weighted scores. The state-based controller adds the local mixture, active disambiguation, and the joint state with adaptive gating. The full controller gains 4 points of SR with VLM logits and 12 with Jev.

\begin{table}[htbp]
\centering\small
\caption{\textbf{Two evaluators crossed with two controllers.} Each row reports 100 closed-loop episodes. Rates are percentages and NE is in meters.}
\label{tab:crossed}
\input{tables/crossed}
\end{table}

\runin{Offline step-wise evaluation}
To isolate a change in the action rule, we hold fixed 1,054 movement decisions from the full Jev trajectories across the 100 episodes. At each step the controllers receive the same questions, options, query targets, and causal observations. Responses are shared within each evaluator. After prediction, we snap candidates to the navigation mesh and identify those nearest the goal by geodesic distance, including ties within $10^{-5}$\,m. An oracle-optimal waypoint hit occurs when the predicted waypoint falls within this set.

\begin{table}[htbp]
\centering\small\setlength{\tabcolsep}{4pt}
\caption{\textbf{Decisions under fixed observations.} Each row evaluates the same 1,054 movement steps. Net hits are relative to score-based control with the same evaluator.}
\label{tab:offline}
\input{tables/offline}
\end{table}

Under Jev, GMM only and local active disambiguation share candidate sets, utilities, and $B_t$. With equal mixture weights and no pairwise evidence, the posterior mean is the arithmetic mean of the assessments. GMM only matches score-based control at 313 oracle-optimal waypoint hits. Restoring the pairwise comparison gives 329 hits. The joint state and adaptive gate give 334, five more on the same decisions.

With high-entropy VLM logits, Eq.~\eqref{eq:gate} retains $B_t$ unless a single viewpoint gives positive worst-case gain. The intermediate variant (GMM and active disambiguation without joint-state gating) reaches 34.0\% SR and 23.4\% SPL on 100 closed-loop episodes. The full controller matches both rates (Table~\ref{tab:crossed}) and gains six oracle-optimal hits over score-based control offline (317 versus 311, Table~\ref{tab:offline}).

\subsection{Deviation Recovery Analysis}
\label{sec:recovery}
We compare BeliefNav with history-based backtracking, a breadth-first-search heuristic on the same topological interface. A deviation begins when the agent first moves more than one graph hop from the mapped reference route. Strict recovery requires returning to a reference node within the next five navigation actions. A return to an earlier route node qualifies.

\begin{figure}[htbp]
\centering
\includegraphics[width=.60\linewidth]{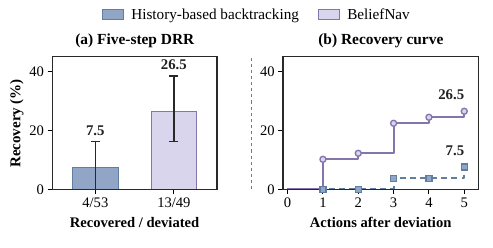}
\caption{\textbf{Return to the reference route after a deviation.} (a) Strict five-action deviation recovery rate (DRR) with 95\% scene-bootstrap intervals. (b) Cumulative strict recovery with the same denominators throughout. Episodes ending before reentry remain unrecovered.}
\label{fig:recovery}
\end{figure}

BeliefNav recovers in 13 of 49 deviated episodes (26.5\%) and backtracking in 4 of 53 (7.5\%). The difference is 19.0 percentage points, with a 95\% paired scene-bootstrap interval of $[5.0,32.0]$ (Fig.~\ref{fig:recovery}). The centered-bootstrap test gives $p=0.0047$, or $0.0188$ after Holm adjustment across the four recovery criteria. The three one-hop intervals include zero (Table~\ref{tab:recovery}). One-hop proximity also counts off-route neighbors across doorways or on another corridor branch, whereas zero-hop reentry identifies a return to the route itself. Appendix~\ref{app:recovery} gives the graph mapping, sampling window, and counts used to resample the ten scenes.

\begin{table}[htbp]
\centering\small\setlength{\tabcolsep}{4pt}
\caption{\textbf{Strict route reentry and one-hop proximity.} Entries are recovered episodes over deviated episodes. Differences and intervals are in percentage points. All four criteria use a five-action window and enter the same Holm correction.}
\label{tab:recovery}
\input{tables/recovery}
\end{table}

\subsection{Failure Modes}
\label{sec:failures}
\begin{figure}[htbp]
\centering
\includegraphics[width=0.94\linewidth]{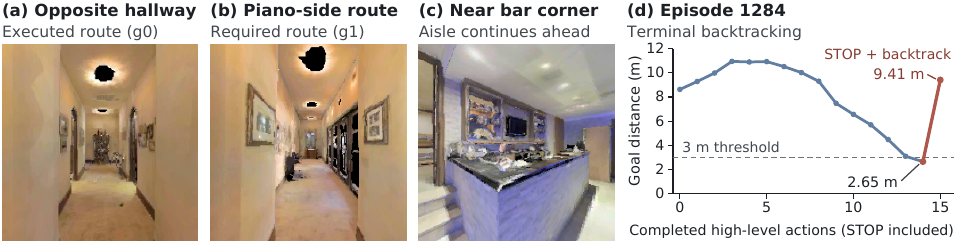}
\caption{\textbf{Visual evidence from the failure audit.} (a,b) The selected hallway and the piano-side alternative in episode 842. (c) The near bar corner and the continuing aisle in episode 7. (d) Goal distance in episode 1284 from the start (0 actions) to terminal backtracking (action 15). DRR measures movement before native \stopact{}.}
\label{fig:failure}
\end{figure}

We reviewed ten episodes sampled without replacement from the 50 full-system failures, assigning one primary category per episode (Table~\ref{tab:failures}). Four meet the criterion for \emph{Shared Judgment Error}, zero for \emph{Persistent False Deviation}, one for \emph{Instruction Ambiguity}, and five for \emph{Execution \& Compound} failure. The last group includes compound cases in which the trace does not isolate one primary mechanism. Figure~\ref{fig:failure} shows three of the reviewed episodes. Appendix~\ref{app:audit} gives the criteria, the utilities for the four shared errors, and the effect of masking historical viewpoints.

\begin{table}[htbp]
\centering\small\setlength{\tabcolsep}{3pt}\renewcommand{\arraystretch}{1.06}
\caption{\textbf{Ten sampled failures grouped by primary category.} NE is the final goal distance. Termination identifies the event that ended the episode.}
\label{tab:failures}
\input{tables/failures}
\end{table}

In episode 842, both assessments favor the hallway opposite the instructed piano-side continuation. In episode 7, the two bar corners support different stopping points. In episode 1284, action 14 (H13) leaves the agent 2.65\,m from the goal, inside the 3\,m success radius. At the 15-action limit (H14), native \stopact{} selects a historical node and backtracks through seven waypoints, ending 9.41\,m from the goal.

\subsection{Comparison with Navigation Policies}
\label{sec:policies}
Table~\ref{tab:policies} places the controlled comparisons in the context of published navigation results. BeliefNav matches the 50.0\% SR of P2DNav with Qwen3-VL-32B. P2DNav-32B reports higher OSR (65.0\% versus 55.0\%) and SPL (30.6\% versus 29.8\%). BeliefNav has lower NE (6.18 versus 6.61\,m). Its SR exceeds the published P2DNav-4B, C\textsuperscript{2}Nav, and NavJev figures by 15, 6, and 23 percentage points, respectively.

\begin{table}[htbp]
\centering\fontsize{8.8}{9.8}\selectfont\renewcommand{\arraystretch}{1.03}
\caption{\textbf{Published R2R-CE validation-unseen results.} Supervised policies and CA-Nav use the full split. The other rows report ON-100. BeliefNav uses ETP-R1's discrete heading interface (Sec.~\ref{sec:setup}). NE is in meters and rates are percentages.}
\label{tab:policies}
\input{tables/policies}
\end{table}

\subsection{Computational Cost and Latency Analysis}
\label{sec:cost}
Across 1,167 high-level decisions, Qwen perception accounts for 94.4\% of mean service time (Table~\ref{tab:cost}). Its 28.7\,s per step reflects sequential, unbatched single-view captioning in the prototype. These instruction-free captions can be batched independently of the CPU belief update. The update runs on 1,067 movement steps and uses 0.172\,s per decision averaged over all steps, including 100 native \stopact{} decisions, with zero additional model requests. Local models run on RTX 4090 GPUs and CPU replay uses one numerical-library thread on an Intel Xeon Gold 6330N. Reported times include Jev API round trips and exclude simulator overhead and system I/O.

\begin{table}[htbp]
\centering\small
\caption{\textbf{Component cost per high-level decision.} A call is a model request for Qwen or Jev and a complete estimator invocation for the CPU row. Service times come from recorded request spans and a single-thread CPU replay. Per-call means average actual invocations. Per-step means include all 1,167 decisions.}
\label{tab:cost}
\input{tables/latency}
\end{table}
\FloatBarrier

%% file: tables/cumulative.tex
\begin{tabular*}{\linewidth}{@{\extracolsep{\fill}}lrrrrrr@{}}
\toprule
Configuration & SR $\uparrow$ & OSR $\uparrow$ & SPL $\uparrow$ & nDTW $\uparrow$ & NE $\downarrow$ & Steps\\
\midrule
Reassessment + historical propagation & 32.0 & 38.0 & 22.7 & 41.1 & 7.20 & 11.68\\
+ Viewpoint weighting & 35.0 & 41.0 & 23.1 & 36.2 & 7.58 & 12.06\\
+ Contextual stage cue & 38.0 & 45.0 & 24.0 & 39.0 & 6.82 & 11.95\\
+ GMM and active disambiguation & 47.0 & 53.0 & 31.1 & 44.8 & 6.00 & 11.38\\
+ Joint state and adaptive gate (full) & \textbf{50.0} & \textbf{55.0} & 29.8 & 41.9 & 6.18 & 11.67\\
\bottomrule
\end{tabular*}

%% file: tables/crossed.tex
\begin{tabular*}{\linewidth}{@{\extracolsep{\fill}}llrrrr@{}}
\toprule
Semantic evaluator & Controller & SR $\uparrow$ & OSR $\uparrow$ & SPL $\uparrow$ & NE $\downarrow$\\
\midrule
VLM logits & Score-based & 30.0 & 42.0 & 18.9 & 7.73\\
VLM logits & State-based & 34.0 & 42.0 & 23.4 & 7.28\\
Jev & Score-based & 38.0 & 45.0 & 24.0 & 6.82\\
Jev & State-based (full) & \textbf{50.0} & \textbf{55.0} & \textbf{29.8} & \textbf{6.18}\\
\bottomrule
\end{tabular*}

%% file: tables/offline.tex
\begin{tabular*}{\linewidth}{@{\extracolsep{\fill}}llrrr@{}}
\toprule
Evaluator & Controller & Oracle-optimal waypoint hits & Rate (\%) & Net hits\\
\midrule
VLM logits & Score-based & 311 & 29.5 & 0\\
VLM logits & Full state-based & 317 & 30.1 & $+6$\\
\addlinespace
Jev & Score-based & 313 & 29.7 & 0\\
Jev & GMM only & 313 & 29.7 & $+0$\\
Jev & GMM + active disambiguation & 329 & 31.2 & $+16$\\
Jev & Full state-based & 334 & 31.7 & $+21$\\
\bottomrule
\end{tabular*}

%% file: tables/recovery.tex
\begin{tabularx}{\linewidth}{@{}Yrrlrr@{}}
\toprule
Recovery criterion & Baseline & BeliefNav & $\Delta$ [95\% CI], pp & $p$ & Holm $p$\\
\midrule
First deviation, strict 0-hop reentry & 4/53 & 13/49 & 19.0 [5.0, 32.0] & 0.0047 & 0.0188\\
\addlinespace
First deviation, 1-hop proximity & 20/53 & 25/49 & 13.3 [$-5.7$, 34.0] & 0.2025 & 0.6074\\
Any deviation onset, 1-hop proximity & 20/53 & 25/49 & 13.3 [$-5.7$, 34.0] & 0.2025 & 0.6074\\
Shared deviated episodes, 1-hop proximity & 15/38 & 19/38 & 10.5 [$-15.4$, 36.0] & 0.4780 & 0.6074\\
\bottomrule
\end{tabularx}

%% file: tables/failures.tex
\begin{tabularx}{\linewidth}{@{}rP{.203\linewidth}rP{.112\linewidth}Y@{}}
\toprule
Episode & Failure category & NE (m) & Termination & Evidence from the trajectory\\
\midrule
362 & Shared Judgment Error & 12.52 & Policy \stopact & Both assessments favor a side exit from the kitchen.\\
842 & Shared Judgment Error & 17.63 & Policy \stopact & Both assessments miss the required piano-side continuation.\\
330 & Shared Judgment Error & 20.06 & Step Limit & Both assessments favor ascent despite an instruction to descend.\\
52 & Shared Judgment Error & 9.14 & Step Limit & Both assessments turn away at the kitchen end.\\
\addlinespace[2pt]
7 & Instruction Ambiguity & 3.66 & Policy \stopact & The near and far bar corners are plausible stopping points.\\
\addlinespace[2pt]
739 & Execution \& Compound & 9.51 & Step Limit & Repeated collisions follow a corrected detour.\\
176 & Execution \& Compound & 4.35 & Step Limit & The assessments disagree, then share an error. Recovery is partial at the step limit.\\
232 & Execution \& Compound & 16.09 & Step Limit & Stair execution, reassessment, pairwise comparison, and viewpoint selection interact.\\
70 & Execution \& Compound & 14.93 & Step Limit & Room grounding is uncertain and the assessments disagree.\\
1284 & Execution \& Compound & 9.41 & Step Limit & Native \stopact{} reverses a within-radius approach.\\
\bottomrule
\end{tabularx}

%% file: tables/policies.tex
\begin{tabular*}{\linewidth}{@{\extracolsep{\fill}}llrrrr@{}}

\toprule
Method & Decision model & NE$\downarrow$ & OSR$\uparrow$ & SR$\uparrow$ & SPL$\uparrow$\\
\midrule
\multicolumn{6}{@{}l}{\textit{Supervised policies (full validation split)}}\\

DUET-CE~\cite{chen2022duet,an2025etpnav} & Transformer & 5.16 & 62.2 & 53.7 & 46.4\\
ETPNav~\cite{an2025etpnav} & Transformer & 4.71 & 65.0 & 57.0 & 49.0\\
BEVBert~\cite{an2023bevbert} & Transformer & 4.57 & 67.0 & 59.0 & 50.0\\
HNR~\cite{wang2024hnr} & HNR & 4.42 & 67.0 & 61.0 & 51.0\\
ETP-R1~\cite{ye2025etpr1} & Graph + GRPO & 3.94 & 72.0 & 65.0 & 56.0\\
\midrule
\multicolumn{6}{@{}l}{\textit{Zero-shot decision models}}\\
Random & -- & 8.63 & 12.0 & 2.0 & 1.5\\
MapGPT-CE~\cite{chen2024mapgpt} & GPT-4o & 8.16 & 21.0 & 7.0 & 5.0\\
DiscussNav-CE~\cite{long2024discuss} & GPT-4 & 7.77 & 15.0 & 11.0 & 10.5\\
Open-Nav~\cite{qiao2025opennav} & GPT-4 & 6.70 & 23.0 & 19.0 & 16.1\\
CA-Nav (full split)~\cite{chen2025canav} & GPT-4 & 7.58 & 48.0 & 25.3 & 10.8\\
SmartWay~\cite{shi2025smartway} & GPT-4o & 7.01 & 51.0 & 29.0 & 22.4\\
P2DNav~\cite{sheng2026p2dnav} & Qwen3-VL-4B & 6.87 & 51.0 & 35.0 & 16.8\\
P2DNav~\cite{sheng2026p2dnav} & Qwen3-VL-32B & 6.61 & \textbf{65.0} & \textbf{50.0} & \textbf{30.6}\\
C\textsuperscript{2}Nav~\cite{zheng2026c2nav} & GPT-5.5 & 7.20 & 54.0 & 44.0 & 29.0\\
NavJev~\cite{sheng2026navjev} & Jev & 7.48 & 35.0 & 27.0 & 22.4\\
\textbf{Ours (full)} & Visual Jev + Jev & \textbf{6.18} & 55.0 & \textbf{50.0} & 29.8\\
\bottomrule
\end{tabular*}

%% file: tables/latency.tex
\begin{tabular*}{\linewidth}{@{\extracolsep{\fill}}lrrr@{}}
\toprule
Layer & Calls per step & Seconds per call & Seconds per step\\
\midrule
Qwen perception and evidence preparation & 16.343 & 1.757 & 28.709\\
Jev semantic decisions (visual and text) & 5.310 & 0.289 & 1.537\\
CPU belief update & 0.914 & 0.188 & 0.172\\
\bottomrule
\end{tabular*}

%% file: sections/discussion.tex
\section{Discussion and Conclusion}
\label{sec:discussion}
\runin{Limitations}
Corrective backtracking can add distance. In the closed-loop comparison, joint estimation raises SR from 47.0\% to 50.0\% while SPL falls from 31.1\% to 29.8\%. Local decisions depend on the semantic evaluator and on lexical matches in the captions, so an error shared by both assessments can persist. Evaluating the method on the full R2R-CE split and on RxR-CE, and revisiting the frozen termination policy, would test how far these results extend.

\runin{Conclusion}
BeliefNav retains disagreement about local waypoints and uncertainty about the stage at a possible wrong turn. A later observation can therefore return control to an earlier viewpoint. On the evaluated subset this improves success and strict route reentry. Deferred commitment provides a lightweight decision layer for zero-shot navigation that can revise a route as the agent moves.

%% file: appendix/settings.tex
\section{Notation and Fixed Inference Settings}
\label{app:settings}
Table~\ref{tab:notation} lists the quantities shared by the movement decision and the auxiliary filters. Table~\ref{tab:settings} records the settings used across the evaluation subset. None was fit on these episodes.

\begin{table}[htbp]
\centering\small
\caption{\textbf{Notation for belief updates and action selection.} A viewpoint can support several active candidates, and a candidate can be visible from several viewpoints.}
\label{tab:notation}
\begin{tabularx}{\linewidth}{@{}lY@{}}
\toprule
Symbol & Meaning\\
\midrule
$p_h^{\rm prop}(v)$ & Historical propagation coefficient from viewpoint $h$ to neighbor $v$.\\
$c_h$ & Incoming propagation mass at viewpoint $h$.\\
$q_t$ & Semantic stage judgment over the $K$ clauses.\\
$z_t$, $C$ & Centered log stage observation and centering matrix $\mathbf I-\mathbf1\mathbf1^\top/K$.\\
$\eta_t$ & Fraction of current observation identities that are new.\\
$e_a,e_b$ & Indicator vectors for candidate coordinates $a,b$ at viewpoint $h$.\\
$d_m$, $v$ & Standardized utility difference and covariance vector for the compared pair in component $m$.\\
$\ell_+,\ell_-$ & Pairwise ranking likelihood factors $q_A+q_U$ and $q_B+q_U$.\\
$Z_m,M_m$ & Unnormalized component mass and first moment after a comparison.\\
$\omega_f$, $\mathcal F_k$ & Stage weight of feature $f$ and the feature set of clause $k$.\\
$\mathcal O$, $\mathcal D_t$ & Caption feature set and the collection eligible at step $t$.\\
$R_t(k),f_t(k)$ & Raw caption support and relative observation factor for stage $k$.\\
$R_{hg}(k),C_{hkg}$ & Candidate-inclusive caption support and the factor replacing viewpoint-only support.\\
$n_h$, $L_h(g)$ & Local candidate set size $|\mathcal G_h|$ and the probability that $g$ wins locally.\\
$B_t^{\rm init},B_t$ & Initial and reassessed viewpoint relevance distributions.\\
$v^D,v^R$ & Belief-informed and prior-only candidate scores.\\
$q^D,q^R$ & Target candidate distributions, uniform on the maximizers of their respective scores.\\
$\mathcal W,\delta_i$ & Admissible viewpoint weights and a one-hot assignment to viewpoint $i$.\\
$A_w(g)$ & Probability that $g$ wins under a utility draw and aggregation weights $w$.\\
$\Delta(w;q)$ & Agreement gain over $B_t$ under target distribution $q$.\\
$\pi_t$ & Normalized historical propagation weights used by the baseline at step $t$.\\
\bottomrule
\end{tabularx}
\end{table}

\begin{table}[htbp]
\centering\small
\caption{\textbf{Fixed settings and their origins.} The minimax gate uses the zero gain of $B_t$ without an additional posterior-mass threshold.}
\label{tab:settings}
\begin{tabularx}{\linewidth}{@{}P{.24\linewidth}Y@{}}
\toprule
Origin & Setting and purpose\\
\midrule
Analytic or neutral reference & $\sigma^2=1/12$ is the variance of $\mathrm{Uniform}(0,1)$. Mixture and transition splits use $1/2$. Neutral stage support is $1/K$ and neutral observations have factor one.\\
Robustness and policy defaults & Student-$t$ degrees of freedom $\nu=4$, log and process noise scales one, and unsupported-coordinate variance four. The query trigger is $0.9$, the odometry tolerance $10^{-3}$\,m, the log floor $10^{-6}$, and the comparison tolerance $10^{-12}$. Historical propagation uses normalized camera projections (Eq.~\eqref{eq:prop}) and has no temperature parameter.\\
Numerical integration & 200,000 draws per viewpoint for local winning probabilities and 32,768 common draws per assignment for action outcomes. Fixed data-derived seeds and separate sampling streams determine the draws.\\
\bottomrule
\end{tabularx}
\end{table}

%% file: appendix/baseline.tex
\section{Baseline and Auxiliary Filter Formulations}
\label{app:baseline}
\runin{Historical propagation}
Initialize $c_t=1$ at the current viewpoint and $c_h=0$ at earlier observations. Process viewpoints from newest to oldest. Let $\mathcal N(h)$ contain all directly connected visited and frontier nodes. For each $u\in\mathcal N(h)$, $s_{hu}$ is the camera score projected onto its bearing as in Eq.~\eqref{eq:projection}. The propagation coefficient to a visited neighbor $v$ is
\begin{equation}
 p_h^{\rm prop}(v)=\frac{s_{hv}}{\sum_{u\in\mathcal N(h)}s_{hu}} .
 \label{eq:prop}
\end{equation}
When the denominator is positive, add $c_h p_h^{\rm prop}(v)$ to the most recent strictly earlier observation at $v$, if one exists. Zero total projection sends no mass. Frontier nodes enter the normalizer, while only visited neighbors pass mass backward. The baseline normalizes the incoming masses $c_h$ over $h\in\mathcal H_t$, the viewpoints with eligible candidates, to obtain $\pi_t$. For each waypoint, Eq.~\eqref{eq:score} uses these weights only at viewpoints that support it.

\runin{Contextual stage cue}
For semantic stage probabilities $q_t$, define the centered log observation
\begin{equation}
 C=\mathbf I-\mathbf1\mathbf1^\top/K,\qquad z_t=C\log\max(q_t,10^{-6}),
 \label{eq:contextlog}
\end{equation}
with the maximum taken elementwise. If all image identities have been seen and support is unchanged, reuse the state. Otherwise let $\eta_t$ be the fraction of new image identities. The random-walk prediction has $m^-=m_{t-1}$ and $P^-=P_{t-1}+Q_t$, where process noise $Q_t$ scales with $\eta_t$ and observation noise $R$ scales inversely with it. For $d=\max(1,K-1)$, use
\begin{equation}
 \varepsilon_t=z_t-m^-,\qquad V=P^-+R,\qquad
 \gamma=\min\!\left(1,\frac{\nu+d}{\nu+\varepsilon_t^\top V^{-1}\varepsilon_t}\right),\quad \nu=4.
 \label{eq:contextrobust}
\end{equation}
Use $R/\gamma$ in the gain and Joseph covariance update, center the updated moments, and average softmax-transformed samples for stage weights. This contextual cue enters semantic assessment. The joint progress--deviation filter in Sec.~\ref{sec:state} has its own state and observation likelihood.

\runin{Raw and corrected departure mass}
For the normalized prediction $\alpha_t^-$, write $m_t^-(a)=\sum_k\alpha_t^-(a,k)$. Historical anchors have unit observation factors. Equation~\eqref{eq:emission} gives the common normalizer and updated raw mass
\begin{align}
 Z_t^{\rm obs}&=\sum_k f_t(k)\alpha_t^-(\mathsf N,k)+\sum_{a\ne\mathsf N}m_t^-(a)
 =1+\sum_k[f_t(k)-1]\alpha_t^-(\mathsf N,k),\label{eq:obsnorm}\\
 m_t(h)&=\sum_k\alpha_t(h,k)=m_t^-(h)/Z_t^{\rm obs}.\label{eq:branchmass}
\end{align}
Support for the predicted normal stages can push $Z_t^{\rm obs}$ above one and reduce every departure mass relative to its prediction. If $Z_t^{\rm obs}<1$, these masses rise. Positive historical branches retain their relative odds during this observation update. Missing or stage-uniform caption support leaves predicted masses unchanged.

On eligible recovery, half the branch mass returns to normal and half stays. For an existing anchor with no new incoming mass, the same transition in the neutral reference cancels in its corrected support before observation normalization,
\begin{equation}
 \frac{\tfrac12\alpha_{t-1}(h,k)}
      {\sum_j\tfrac12\alpha_{t-1}^0(h,j)}=\lambda_{t-1}(h,k).
 \label{eq:cancellation}
\end{equation}
Subsequent evidence satisfying $\sum_k[f_t(k)-1]\alpha_t^-(\mathsf N,k)>0$ reduces the retained raw departure mass through Eq.~\eqref{eq:branchmass}. Raw mass, corrected support, and the current viewpoint aggregation weight therefore need not change together.

\runin{Observation inputs and state persistence}
The original instruction and its ordered verbatim clauses remain available. Reassessment sees the observed camera views and full instruction. The initial assessment can use a fallible stage hint, and initial viewpoint relevance can use the stage distribution. Reestimated relevance compares available continuations using the full instruction, executed history, candidate-specific descriptions, and an unresolved option. Pairwise queries use the associated camera views, guide coordinates, and estimated waypoint distances. The observer receives no navigation instruction or candidate scores.

Evidence is indexed by decision step. The system stores $B_t$ before gating, so derived weights cannot reenter as new observations. A change to the underlying utility assessment resets the old pairwise factor. A change in available candidates alone can preserve archived pair variables for a valid marginal. Inactive recovery anchors persist across changes to the active candidate set.

%% file: appendix/recovery.tex
\section{Statistical Details of Deviation Recovery}
\label{app:recovery}
\runin{Mapping a trajectory to the route}
The original Matterport3D connectivity graph $G_i$ contains the included viewpoints and unit-cost, mutually unobstructed edges. The original trajectory identifiers give the set $P_i$ of reference nodes. Let $\pi_i(x)$ map a pose to its nearest viewpoint in three-dimensional camera coordinates after the 1.25\,m sensor-height offset. For episode $i$, method $m$, and pose $x_{imt}$, the hop distance is
\begin{equation}
 d_{imt}=\min_{v\in P_i}\operatorname{dist}_{G_i}\big(\pi_i(x_{imt}),v\big).
 \label{eq:hops}
\end{equation}
Zero hops denotes membership in $P_i$ after mapping. The largest camera-to-viewpoint residuals are 2.62\,m for BeliefNav and 3.51\,m for backtracking.

\runin{Window and denominator}
We sample the initial pose and the end of each executed high-level movement. Several low-level motions, collisions, or backtracking waypoints within an action count once. For the first deviation $\tau_{im}=\min\{t:d_{imt}>1\}$, let $R_{im}(5)=1$ if an observed action endpoint reaches $d_{im,\tau_{im}+j}=0$ for some $1\le j\le5$. The departure action and native \stopact{}, including terminal backtracking, are excluded. An episode that ends before recovery remains in the denominator. A later departure does not undo a recorded reentry. With $D_m$ the set of episodes that deviate under method $m$,
\begin{equation}
 \widehat{\mathrm{DRR}}_m=\frac{\sum_{i\in D_m}R_{im}(5)}{|D_m|}.
 \label{eq:drr}
\end{equation}
Of the 100 episodes, 38 deviate under both methods, 15 under backtracking alone, 11 under BeliefNav alone, and 36 under neither. The primary comparison therefore conditions on each method's own deviated episodes. The shared-episode row in Table~\ref{tab:recovery} uses one-hop proximity.

\runin{Paired scene bootstrap}
We sample ten scenes with replacement, using the same scene multiplicities for both methods, and compute each rate from the pooled counts in Table~\ref{tab:scenes}. Scenes with no deviations remain eligible. All 100,000 resamples (seed 20261117) give positive pooled denominators. The 2.5th and 97.5th percentiles of the paired differences give the 95\% interval. For observed difference $\widehat\Delta$ and centered replicate $\delta_b=\Delta_b^*-\widehat\Delta$, the two-sided test uses
\begin{equation}
 p=\frac{1+\sum_{b=1}^{B}\mathbf1\{|\delta_b|\ge|\widehat\Delta|\}}{B+1},
 \qquad B=100{,}000 .
 \label{eq:bootstrap}
\end{equation}
The strict criterion has 469 exceedances. Table~\ref{tab:recovery} also reports Holm-adjusted values for the four recovery criteria. Inference uses the ten observed scene clusters.

\begin{table}[htbp]
\centering\small
\caption{\textbf{Strict recoveries by scene.} Each cell gives recovered over deviated episodes. A zero denominator indicates a scene with no deviated episode for that method.}
\label{tab:scenes}
\input{tables/scenes}
\end{table}

%% file: tables/scenes.tex
\begin{tabular*}{.72\linewidth}{@{\extracolsep{\fill}}lrr@{}}
\toprule
Scene & History-based backtracking & BeliefNav \\
\midrule
2azQ1b91cZZ & 0/9 & 2/11 \\
8194nk5LbLH & 0/0 & 0/0 \\
EU6Fwq7SyZv & 1/3 & 2/3 \\
QUCTc6BB5sX & 2/10 & 1/8 \\
TbHJrupSAjP & 1/6 & 2/5 \\
X7HyMhZNoso & 0/3 & 0/2 \\
Z6MFQCViBuw & 0/5 & 1/6 \\
oLBMNvg9in8 & 0/3 & 1/2 \\
x8F5xyUWy9e & 0/0 & 0/2 \\
zsNo4HB9uLZ & 0/14 & 4/10 \\
\bottomrule
\end{tabular*}

%% file: appendix/audit.tex
\section{Detailed Failure Audit Diagnostics}
\label{app:audit}
\runin{Sampling and attribution}
We sorted the 50 full-system failure IDs and sampled ten without replacement using seed 20261121. The sample and criteria were fixed before detailed inspection. The authors reviewed each of the ten trajectories, aligning its instruction and executed motions with the images, local utilities, pairwise comparisons, and belief history. The audit included checks of image identity and numerical consistency, with AI assistance in the visual inspection of critical views. Each episode receives one primary category for its earliest substantiated unresolved failure chain.

\begin{description}[leftmargin=1.25em,itemsep=3pt]
 \item[Shared Judgment Error.] A valid continuation is visible, both assessments prefer an incompatible continuation at that viewpoint, and this preference persists after any recorded pairwise update along the failure chain.
 \item[Persistent False Deviation.] An identifiable branch starts at an ambiguous motion, later evidence contradicts it, and it continues to affect the failing decisions. Retaining positive latent mass alone is insufficient.
 \item[Instruction Ambiguity.] Two interpretations of the complete instruction are consistent with the scene and executed history and imply different continuations or stopping points.
 \item[Execution \& Compound.] Interacting semantic judgments, execution, or termination effects prevent a single-cause assignment. This category also includes cases with insufficient evidence to isolate a primary cause.
\end{description}

\runin{Critical local decisions}
In the four shared-error cases, both original and reassessed utilities favor the executed error over the reviewed alternative (Table~\ref{tab:utilities}). The selected viewpoint has unit weight, and there is no pairwise update at these steps. In episode 330 the assessments initially disagree, then both favor ascent at H3 despite the instruction to descend. In episode 52, corrective actions follow earlier detours, but at H12 both assessments favor moving away from the kitchen-end continuation.

\begin{table}[htbp]
\centering\small
\caption{\textbf{Local utilities at the four shared judgment errors.} Within the selected viewpoint, $e$ is the executed error and $a$ the reviewed alternative. The columns give the original scores $s$, reassessed scores $r$, and posterior means $\mu$. Step is the high-level decision index.}
\label{tab:utilities}
\input{tables/utilities}
\end{table}

\runin{Masking a historical viewpoint}
To test whether an earlier viewpoint affects the current choice, we mask its row in $J_t$, renormalize, and recompute the belief-informed target distribution $q^D$ and the gate. The prior-only target distribution $q^R$, utility posteriors, action outcomes, and admissible assignments stay fixed. Among the audited cases only episode 52 changes a maximizing candidate. Masking viewpoint 4 changes H5 from g11 to g12 and H6 from g9 to g12. Masking viewpoint 8 changes H9 from g19 to g18. The recorded H6 and H9 moves were corrective. The later shared ranking error at H12 supports this episode's primary attribution.

In episode 739 a corrected detour precedes repeated collisions and terminal stagnation. In episode 1284 the terminal displacement itself is observed (Fig.~\ref{fig:failure}d). For episodes 176, 232, and 70, the traces combine disagreement, incomplete recovery, or execution effects without resolving their relative contributions.

%% file: tables/utilities.tex
\begin{tabular*}{\linewidth}{@{\extracolsep{\fill}}rrllrrrrrr@{}}
\toprule
Ep. & Step & Error & Alt. & $s_e$ & $s_a$ & $r_e$ & $r_a$ & $\mu_e$ & $\mu_a$ \\
\midrule
362 & 0 & g2 & g0 & 0.501 & 0.109 & 0.744 & 0.247 & 0.622 & 0.178 \\
842 & 0 & g0 & g1 & 0.793 & 0.032 & 0.855 & 0.055 & 0.824 & 0.043 \\
330 & 3 & g10 & g8 & 0.639 & 0.202 & 0.887 & 0.095 & 0.763 & 0.149 \\
52 & 12 & g26 & g27 & 0.703 & 0.083 & 0.880 & 0.051 & 0.791 & 0.067 \\
\bottomrule
\end{tabular*}

%% file: main.bbl
\begin{thebibliography}{36}
\providecommand{\natexlab}[1]{#1}
\providecommand{\url}[1]{\texttt{#1}}
\expandafter\ifx\csname urlstyle\endcsname\relax
  \providecommand{\doi}[1]{doi: #1}\else
  \providecommand{\doi}{doi: \begingroup \urlstyle{rm}\Url}\fi

\bibitem[An et~al.(2023)An, Qi, Li, Huang, Wang, Tan, and Shao]{an2023bevbert}
Dong An, Yuankai Qi, Yangguang Li, Yan Huang, Liang Wang, Tieniu Tan, and Jing
  Shao.
\newblock {BEVBert}: Multimodal map pre-training for language-guided
  navigation.
\newblock In \emph{Proceedings of the IEEE/CVF International Conference on
  Computer Vision}, 2023.

\bibitem[An et~al.(2025)An, Wang, Wang, Wang, Huang, He, and
  Wang]{an2025etpnav}
Dong An, Hanqing Wang, Wenguan Wang, Zun Wang, Yan Huang, Keji He, and Liang
  Wang.
\newblock {ETPNav}: Evolving topological planning for vision-language
  navigation in continuous environments.
\newblock \emph{IEEE Transactions on Pattern Analysis and Machine
  Intelligence}, 47\penalty0 (7):\penalty0 5130--5145, 2025.

\bibitem[Anderson et~al.(2018)Anderson, Wu, Teney, Bruce, Johnson,
  S{\"u}nderhauf, Reid, Gould, and van~den Hengel]{anderson2018vln}
Peter Anderson, Qi Wu, Damien Teney, Jake Bruce, Mark Johnson, Niko
  S{\"u}nderhauf, Ian Reid, Stephen Gould, and Anton van~den Hengel.
\newblock Vision-and-language navigation: Interpreting visually-grounded
  navigation instructions in real environments.
\newblock In \emph{Proceedings of the IEEE Conference on Computer Vision and
  Pattern Recognition}, 2018.

\bibitem[Bai et~al.(2025{\natexlab{a}})Bai, Cai, Chen, et~al.]{bai2025qwen3vl}
Shuai Bai, Yuxuan Cai, Ruizhe Chen, et~al.
\newblock {Qwen3-VL} technical report.
\newblock \emph{arXiv preprint arXiv:2511.21631}, 2025{\natexlab{a}}.

\bibitem[Bai et~al.(2025{\natexlab{b}})Bai, Chen, Liu, et~al.]{bai2025qwen25vl}
Shuai Bai, Keqin Chen, Xuejing Liu, et~al.
\newblock {Qwen2.5-VL} technical report.
\newblock \emph{arXiv preprint arXiv:2502.13923}, 2025{\natexlab{b}}.

\bibitem[Bajcsy(1988)]{bajcsy1988active}
Ruzena Bajcsy.
\newblock Active perception.
\newblock \emph{Proceedings of the IEEE}, 76\penalty0 (8):\penalty0 996--1005,
  1988.

\bibitem[Bishop(2006)]{bishop2006prml}
Christopher~M. Bishop.
\newblock \emph{Pattern Recognition and Machine Learning}.
\newblock Springer, 2006.

\bibitem[Chen et~al.(2024{\natexlab{a}})Chen, Lin, Xu, Chai, Liang, and
  Wong]{chen2024mapgpt}
Jiaqi Chen, Bingqian Lin, Ran Xu, Zhenhua Chai, Xiaodan Liang, and Kwan-Yee
  Wong.
\newblock {MapGPT}: Map-guided prompting with adaptive path planning for
  vision-and-language navigation.
\newblock In \emph{Proceedings of the 62nd Annual Meeting of the Association
  for Computational Linguistics}, pages 9796--9810, 2024{\natexlab{a}}.

\bibitem[Chen et~al.(2024{\natexlab{b}})Chen, An, Huang, Xu, Su, Ling, Reid,
  and Wang]{chen2025canav}
Kehan Chen, Dong An, Yan Huang, Rongtao Xu, Yifei Su, Yonggen Ling, Ian Reid,
  and Liang Wang.
\newblock Constraint-aware zero-shot vision-language navigation in continuous
  environments.
\newblock \emph{arXiv preprint arXiv:2412.10137}, 2024{\natexlab{b}}.

\bibitem[Chen et~al.(2021)Chen, Guhur, Schmid, and Laptev]{chen2021hamt}
Shizhe Chen, Pierre-Louis Guhur, Cordelia Schmid, and Ivan Laptev.
\newblock History aware multimodal transformer for vision-and-language
  navigation.
\newblock In \emph{Advances in Neural Information Processing Systems}, 2021.

\bibitem[Chen et~al.(2022)Chen, Guhur, Tapaswi, Schmid, and
  Laptev]{chen2022duet}
Shizhe Chen, Pierre-Louis Guhur, Makarand Tapaswi, Cordelia Schmid, and Ivan
  Laptev.
\newblock Think global, act local: Dual-scale graph transformer for
  vision-and-language navigation.
\newblock In \emph{Proceedings of the IEEE/CVF Conference on Computer Vision
  and Pattern Recognition}, pages 16537--16547, 2022.

\bibitem[Fox et~al.(1999)Fox, Burgard, and Thrun]{fox1999markov}
Dieter Fox, Wolfram Burgard, and Sebastian Thrun.
\newblock Markov localization for mobile robots in dynamic environments.
\newblock \emph{Journal of Artificial Intelligence Research}, 11:\penalty0
  391--427, 1999.

\bibitem[Ilharco et~al.(2019)Ilharco, Jain, Ku, Ie, and
  Baldridge]{ilharco2019ndtw}
Gabriel Ilharco, Vihan Jain, Alexander Ku, Eugene Ie, and Jason Baldridge.
\newblock General evaluation for instruction conditioned navigation using
  dynamic time warping.
\newblock \emph{arXiv preprint arXiv:1907.05446}, 2019.

\bibitem[Jiang et~al.(2026)Jiang, Li, and Li]{jiang2026memory}
Dongming Jiang, Yi Li, and Bingzhe Li.
\newblock {Jev-Mem}: System-one-controlled agentic memory for efficient {AI}
  agents.
\newblock \emph{arXiv preprint arXiv:2609.23986}, 2026.

\bibitem[Krantz et~al.(2020)Krantz, Wijmans, Majumdar, Batra, and
  Lee]{krantz2020vlnce}
Jacob Krantz, Erik Wijmans, Arjun Majumdar, Dhruv Batra, and Stefan Lee.
\newblock Beyond the nav-graph: Vision-and-language navigation in continuous
  environments.
\newblock In \emph{European Conference on Computer Vision}, 2020.

\bibitem[Krantz et~al.(2021)Krantz, Gokaslan, Batra, Lee, and
  Maksymets]{krantz2021waypoint}
Jacob Krantz, Aaron Gokaslan, Dhruv Batra, Stefan Lee, and Oleksandr Maksymets.
\newblock Waypoint models for instruction-guided navigation in continuous
  environments.
\newblock In \emph{Proceedings of the IEEE/CVF International Conference on
  Computer Vision}, pages 15162--15171, 2021.

\bibitem[Li et~al.(2026{\natexlab{a}})Li, Wang, Gong, Lang, and Yu]{li2026edge}
Delong Li, Xu Wang, Haochen Gong, Rui Lang, and Guangsheng Yu.
\newblock Replacing large language models with {Jev} decision models for
  low-latency edge service orchestration.
\newblock \emph{arXiv preprint arXiv:2609.22753}, 2026{\natexlab{a}}.
\newblock Version 2.

\bibitem[Li et~al.(2026{\natexlab{b}})Li, Miao, Krishnan, and
  Padman]{li2026judge}
Yubo Li, Yidi Miao, Ramayya Krishnan, and Rema Padman.
\newblock {JEV-as-a-Judge}: Accept when confident, escalate when unsure.
\newblock \emph{arXiv preprint arXiv:2609.26550}, 2026{\natexlab{b}}.
\newblock Version 5.

\bibitem[Long et~al.(2023)Long, Li, Cai, and Dong]{long2024discuss}
Yuxing Long, Xiaoqi Li, Wenzhe Cai, and Hao Dong.
\newblock Discuss before moving: Visual language navigation via multi-expert
  discussions.
\newblock \emph{arXiv preprint arXiv:2309.11382}, 2023.

\bibitem[Long et~al.(2024)Long, Cai, Wang, Zhan, and Dong]{long2024instructnav}
Yuxing Long, Wenzhe Cai, Hongcheng Wang, Guanqi Zhan, and Hao Dong.
\newblock {InstructNav}: Zero-shot system for generic instruction navigation in
  unexplored environment.
\newblock \emph{arXiv preprint arXiv:2406.04882}, 2024.

\bibitem[Ma et~al.(2019{\natexlab{a}})Ma, Lu, Wu, AlRegib, Kira, Socher, and
  Xiong]{ma2019selfmonitoring}
Chih-Yao Ma, Jiasen Lu, Zuxuan Wu, Ghassan AlRegib, Zsolt Kira, Richard Socher,
  and Caiming Xiong.
\newblock Self-monitoring navigation agent via auxiliary progress estimation.
\newblock In \emph{International Conference on Learning Representations},
  2019{\natexlab{a}}.

\bibitem[Ma et~al.(2019{\natexlab{b}})Ma, Wu, AlRegib, Xiong, and
  Kira]{ma2019regretful}
Chih-Yao Ma, Zuxuan Wu, Ghassan AlRegib, Caiming Xiong, and Zsolt Kira.
\newblock The regretful agent: Heuristic-aided navigation through progress
  estimation.
\newblock In \emph{Proceedings of the IEEE/CVF Conference on Computer Vision
  and Pattern Recognition}, pages 6732--6740, 2019{\natexlab{b}}.

\bibitem[Peng et~al.(2026)Peng, Guo, Xu, Liu, Yan, Ye, Li, and
  Wang]{peng2026dgnav}
Jiankun Peng, Jianyuan Guo, Ying Xu, Yue Liu, Jiashuang Yan, Xuanwei Ye, Houhua
  Li, and Xiaoming Wang.
\newblock Dynamic topology awareness: Breaking the granularity rigidity in
  vision-language navigation.
\newblock \emph{arXiv preprint arXiv:2601.21751}, 2026.

\bibitem[Qiao et~al.(2025)Qiao, Lyu, Wang, Wang, Li, Zhang, Tan, and
  Wu]{qiao2025opennav}
Yanyuan Qiao, Wenqi Lyu, Hui Wang, Zixu Wang, Zerui Li, Yuan Zhang, Mingkui
  Tan, and Qi Wu.
\newblock {Open-Nav}: Exploring zero-shot vision-and-language navigation in
  continuous environment with open-source {LLMs}.
\newblock In \emph{IEEE International Conference on Robotics and Automation},
  2025.

\bibitem[Sheng et~al.(2026{\natexlab{a}})Sheng, Wang, Dai, Li, Qin, He, Liu,
  and Chen]{sheng2026p2dnav}
Kai Sheng, Liuyi Wang, Haojie Dai, Jinlong Li, Yongrui Qin, Zongtao He, Chengju
  Liu, and Qijun Chen.
\newblock {P2DNav}: Panorama-to-downview reasoning for zero-shot
  vision-and-language navigation.
\newblock \emph{arXiv preprint arXiv:2605.19634}, 2026{\natexlab{a}}.

\bibitem[Sheng et~al.(2026{\natexlab{b}})Sheng, Wang, Li, Dai, Liu, and
  Chen]{sheng2026navjev}
Kai Sheng, Liuyi Wang, Jinlong Li, Haojie Dai, Chengju Liu, and Qijun Chen.
\newblock {NavJev}: Efficient vision-language navigation via action-centric
  visual compression and discriminative action-semantic memory.
\newblock \emph{arXiv preprint arXiv:2609.34969}, 2026{\natexlab{b}}.

\bibitem[Shi et~al.(2025)Shi, Li, Lyu, Xia, Dayoub, Qiao, and
  Wu]{shi2025smartway}
Xiangyu Shi, Zerui Li, Wenqi Lyu, Jiatong Xia, Feras Dayoub, Yanyuan Qiao, and
  Qi Wu.
\newblock {SmartWay}: Enhanced waypoint prediction and backtracking for
  zero-shot vision-and-language navigation.
\newblock In \emph{IEEE/RSJ International Conference on Intelligent Robots and
  Systems}, 2025.

\bibitem[Thrun et~al.(2005)Thrun, Burgard, and Fox]{thrun2005probabilistic}
Sebastian Thrun, Wolfram Burgard, and Dieter Fox.
\newblock \emph{Probabilistic Robotics}.
\newblock MIT Press, 2005.

\bibitem[{TypeSafe AI}(n.d.)]{typesafe2026api}
{TypeSafe AI}.
\newblock {System One}.
\newblock \url{https://docs.typesafe.ai/api/system-one}, n.d.
\newblock API reference. Accessed November 17, 2026.

\bibitem[Wang et~al.(2021)Wang, Wang, Liang, Xiong, and Shen]{wang2021ssm}
Hanqing Wang, Wenguan Wang, Wei Liang, Caiming Xiong, and Jianbing Shen.
\newblock Structured scene memory for vision-language navigation.
\newblock In \emph{Proceedings of the IEEE/CVF Conference on Computer Vision
  and Pattern Recognition}, pages 8455--8464, 2021.

\bibitem[Wang et~al.(2024)Wang, Li, Yang, Liu, Hu, Jiang, and
  Jiang]{wang2024hnr}
Zihan Wang, Xiangyang Li, Jiahao Yang, Yeqi Liu, Junjie Hu, Ming Jiang, and
  Shuqiang Jiang.
\newblock Lookahead exploration with neural radiance representation for
  continuous vision-language navigation.
\newblock In \emph{Proceedings of the IEEE/CVF Conference on Computer Vision
  and Pattern Recognition}, pages 13753--13762, 2024.

\bibitem[Yao et~al.(2025)Yao, Gao, and Xu]{yao2025navmorph}
Xuan Yao, Junyu Gao, and Changsheng Xu.
\newblock {NavMorph}: A self-evolving world model for vision-and-language
  navigation in continuous environments.
\newblock In \emph{Proceedings of the IEEE/CVF International Conference on
  Computer Vision}, pages 5536--5546, 2025.

\bibitem[Ye et~al.(2025)Ye, Mao, Cui, Yu, Zhai, Chen, Zhou, Xiong, and
  Wang]{ye2025etpr1}
Shuhao Ye, Sitong Mao, Yuxiang Cui, Xuan Yu, Shichao Zhai, Wen Chen, Shunbo
  Zhou, Rong Xiong, and Yue Wang.
\newblock {ETP-R1}: Evolving topological planning with reinforcement
  fine-tuning for vision-language navigation in continuous environments.
\newblock \emph{arXiv preprint arXiv:2512.20940}, 2025.

\bibitem[Yu and Yao(2026)]{yu2026visual}
Guanxu Yu and Yuhang Yao.
\newblock Visual {Jev}: Accurate and efficient decisions from shared visual
  context.
\newblock \emph{arXiv preprint arXiv:2609.25845}, 2026.

\bibitem[Zheng et~al.(2026)Zheng, Zhang, and Liu]{zheng2026c2nav}
Runtian Zheng, Congpeng Zhang, and Ying Liu.
\newblock {C\textsuperscript{2}Nav}: Compare before you commit for zero-shot
  vision-and-language navigation.
\newblock \emph{arXiv preprint arXiv:2609.15142}, 2026.

\bibitem[Zhou et~al.(2024)Zhou, Hong, and Wu]{zhou2024navgpt}
Gengze Zhou, Yicong Hong, and Qi Wu.
\newblock {NavGPT}: Explicit reasoning in vision-and-language navigation with
  large language models.
\newblock In \emph{Proceedings of the AAAI Conference on Artificial
  Intelligence}, 2024.

\end{thebibliography}
